\documentclass[journal]{IEEEtran}

\ifCLASSINFOpdf
\else
   \usepackage[dvips]{graphicx}
\fi
\usepackage{url}

\ifCLASSOPTIONcompsoc
    \usepackage[caption=false, font=normalsize, labelfont=sf, textfont=sf]{subfig}
\else
\usepackage[caption=false, font=footnotesize]{subfig}
\fi

\usepackage{graphicx, amsmath, amssymb, amsthm, bm}
\usepackage{hyperref}
\usepackage{algorithm, algorithmic} 
\usepackage{xcolor}

\newcommand{\lp}{\left(}
\newcommand{\rp}{\right)}
\newcommand{\lc}{\left\{}
\newcommand{\rc}{\right\}}

\newcommand{\R}{\mathbb{R}}
\newcommand{\calL}{\mathcal{L}}
\newcommand{\calS}{\mathcal{S}}

\newcommand\expec[1]{\mathbb{E}\left[ #1 \right]}
\DeclareMathOperator{\sign}{sign}

\newtheorem{thm}{Theorem}[]
\newtheorem{lemma}{Lemma}[]
\newtheorem{assumption}{Assumption}[]

\begin{document}

\title{Gossiped and Quantized Online Multi-Kernel Learning}

\author{Tomas Ortega, \IEEEmembership{Student Member, IEEE}, and Hamid Jafarkhani, \IEEEmembership{Fellow, IEEE}
   \thanks{
   Authors are with the Center for Pervasive Communications \& Computing and  EECS Department, University of California, Irvine, Irvine, CA 92697 USA (e-mail: \{tomaso, hamidj\}@uci.edu). This work was supported in part by the NSF Award ECCS-2207457.} }

\markboth{DOI: 10.1109/LSP.2023.3268988}{}

\maketitle

\begin{abstract}
   In instances of online kernel learning where little prior information is available and centralized learning is unfeasible, past research has shown that distributed and online multi-kernel learning provides sub-linear regret as long as every pair of nodes in the network can communicate (i.e., the communications network is a complete graph).
   In addition, to manage the communication load, which is often a performance bottleneck, communications between nodes can be quantized.
   This letter expands on these results to non-fully connected graphs, which is often the case in wireless sensor networks.
   To address this challenge, we propose a gossip algorithm and provide a proof that it achieves sub-linear regret.
   Experiments with real datasets confirm our findings.
\end{abstract}


\IEEEpeerreviewmaketitle

\section{Introduction}

\IEEEPARstart{M}{any} of the latest efforts in machine learning are focused on bringing learning as close to data collection as possible.
This is of practical interest in a diverse array of applications, sensor networks in particular.
In sensor networks, nodes are often communication-constrained, operate distributively, and can send only a few bits to their neighbors.

There are several papers in the literature that consider the problem of designing distributed learning methods for Online Multi-Kernel Learning (OMKL).
In particular, \cite{OMKL} contains the design and regret analysis of quantized and distributed OMKL for a fully connected (complete) graph.
Although the fully connected case is important, it is not applicable in many sensor networks.
This letter designs a new algorithm that expands the theoretical guarantees of the distributed and quantized OMKL in \cite{OMKL} to an arbitrary general non-fully connected graph.
Unlike \cite{OMKL}, our method uses a novel hidden state quantization scheme in conjunction with a gossip algorithm. One major advantage, compared with \cite{OMKL}, is that our algorithm does not need to communicate the loss function.
Another example of a distributed OMKL that works for a non-fully connected graph is presented in \cite{hong2021distributed}. It requires communicating an unlimited number of bits between neighbors, and its assumptions are stricter than those of \cite{OMKL}.
To manage the required communication throughput, which is often a performance bottleneck \cite{savazzi2021opportunities, kairouz2021advances}, this letter provides the properties of \cite{hong2021distributed} under looser assumptions and while communicating only a limited number of bits between neighboring nodes.

In \cite{Koloskova2019}, a distributed stochastic gradient descent scheme with quantized communication is discussed.
The proposed algorithm uses an extra shared ``hidden'' state and can be used for the single-kernel case, but not multi-kernel settings.
In addition to extending \cite{OMKL} to general non-fully connected networks, our work can be considered as the extension of \cite{Koloskova2019} to multi-kernel cases.

There are a variety of learning tasks that OMKL is suited for, such as multi-kernel classification, regression, and clustering \cite{OMKC, omkr, unified_clustering}.
For a more in-depth review of the literature, see \cite{OMKL}.

\section{Problem formulation}

Let us consider a network with $J$ nodes and model it as an undirected connected graph, where two nodes are connected if and only if they can communicate.
Our set of nodes $V$, and the set of edges $E$, define a communication graph $G=(V,E)$.
At each instant of time $t$, Node $j$ receives the data string $\bm{x}^j_t \in \R^d$ and the desired response $y^j_t \in \R$.
Our task is to find an approximating function $f(\bm{x}^j_t)$ for $y^j_t$.
As in \cite{OMKL}, 
$f$ belongs to a Reproducing Kernel Hilbert Space (RKHS) $\mathcal{H} = \lc f \mid f(\bm{x}) = \sum_{t=1}^\infty \sum_{j=1}^J \alpha^j_t \kappa (\bm{x}, \bm{x}^j_t) \rc$, where $\kappa(\bm{x}, \bm{x}^j_t) : \R^d \times \R^d \to \R$ is a kernel function that measures the similarity between $x$ and $\bm{x}^j_t$.
Let us consider the following optimization problem
\begin{equation} \label{eq:original_problem}
   \min_{f\in \mathcal{H}} \sum_{t=1}^{T} \sum_{j=1}^J \mathcal{C} \lp f(\bm{x}^j_t), y^j_t\rp + \lambda \Omega \lp \lVert f \rVert_\mathcal{H}^2 \rp,
\end{equation}
where $\mathcal{C}$ is a cost function and $\lambda > 0$ is a regularization parameter that controls an increasing function $\Omega$.
An optimal solution for this problem exists in the form $\hat f(\bm{x}) = \sum_{t=1}^{T} \sum_{j=1}^J \alpha_t^j \kappa (\bm{x}, \bm{x}_t^j )$, where $\alpha_t^j$ are real numbers \cite{scholkopf2002learning}.

In multi-kernel learning, a weighted combination of several kernels is selected to improve the performance, compared to single-kernel learning.
A convex combination of the kernels $\lc \kappa_p \rc_{p=1}^P$, where $\kappa_p \in \mathcal{H}_p$ is an RKHS, is also an RKHS, denoted by $\bar{\mathcal{H}} = \mathcal{H}_1 \oplus \cdots \oplus \mathcal{H}_P$ \cite{scholkopf2002learning}.
Here, $\oplus$ indicates the direct sum of Hilbert spaces.
Using $\bar{\mathcal{H}}$ instead of $\mathcal{H}$, we turn problem \eqref{eq:original_problem} into:
\begin{subequations}
   \begin{multline}
      \min_{ \lc \bar w_p^j \rc, \lc f_p \rc } \sum_{t=1}^{T} \sum_{j=1}^J \mathcal{C} \lp \sum_{p=1}^P \bar{w}_p^j f_p(\bm{x}_t^j), y_t^j \rp + \\
      \lambda \Omega \lp \lVert \sum_{p=1}^P \bar w_p^j f_p \rVert^2_{\bar{\mathcal{H}}} \rp,
   \end{multline}
   \begin{equation}
      \text{s.t. } \sum_{p=1}^P \bar w_p^j = 1, \, \bar w_p^j \geq 0, f_p \in \mathcal{H}_p. 
   \end{equation}
\end{subequations}

To evade the curse of dimensionality, a Random Feature (RF) approximation is used \cite{OMKL, Shen2019}.
For normalized shift-invariant kernels, i.e., $\kappa(\bm{x}, \bm{x}') = \kappa (\bm{x} - \bm{x}')$, their Fourier transforms $\pi_\kappa(v)$ exist, and due to normalization, $\pi_\kappa (0) = 1$.
Also, viewing $\pi_\kappa$ as a pdf, $\kappa(\bm{x} - \bm{x}') = \int \pi_\kappa (v) e^{j v^\top (\bm{x} - \bm{x}')} \, dv = \mathbb{E}_v [ e^{jv^\top (\bm{x}-\bm{x}')} ]$.
Gaussian, Laplacian, and Cauchy kernels are three examples of normalized shift-invariant kernels that we can use \cite{random_features}. However, any family of shift invariant kernels can be used for the RF approximation we describe next.
Drawing $D$ i.i.d. samples $\{\bm{v_i}\}_{i=1}^D$ from $\pi_\kappa(v)$, we define
\begin{equation}
    \begin{aligned}
    z_V(\bm{x}) = \frac{1}{\sqrt{D}} [&\sin(\bm{v_1^\top x}), \ldots, \sin(\bm{v_D^\top x}), \\
    &\cos(\bm{v_1^\top x}), \ldots, \cos(\bm{v_D^\top x})]^\top, 
    \end{aligned}
\end{equation}
which has the property that $\kappa(\bm{x}, \bm{x'}) = \mathbb{E}_v [z_V(\bm{x'})^\top z_V(\bm{x})]$.
So, given a fixed $\{\bm{v_i}\}_{i=1}^D$, we can approximate $\kappa(\bm{x}, \bm{x'}) \approx z_V(\bm{x'})^\top z_V(\bm{x})$.
This is called the RF approximation, and it is an unbiased and consistent estimation of $\kappa(\bm{x}, \bm{x'})$ \cite{Shen2019}.
Thus, a weight vector $\bm{\theta} \in \R^{2D}$ can be constructed such that
\begin{multline}
    \hat f(\bm{x}) = \sum_{t=1}^{T} \sum_{j=1}^J \alpha_t^j \kappa (\bm{x}, \bm{x}_t^j ) \approx \sum_{t=1}^{T} \sum_{j=1}^J \alpha_t^j z_V(\bm{x}_t^j)^\top z_V(\bm{x}) \\
    =\bm{\theta}^\top z_V(\bm{x}).
\end{multline}
Then, the loss function is defined as
\begin{equation} \label{eq:loss_function}
   \calL(f(\bm{x})) 
   = \mathcal{C} (\bm{\theta}^\top z_V(\bm{x}), y) + \lambda \Omega (\lVert \bm{\theta} \rVert^2),
\end{equation}
and the following equations are obtained for Kernel $p$ and Node $j$:
\begin{align}
   \hat f^j_{p,t} (\bm{x}_t^j) & = {\bm{\theta}^j_{p,t}}^\top z_{V_p}(\bm{x}_t^j),                                                                          \\
   \bm{\theta}_{p, t+1}^j      & = \bm{\theta}_{p,t}^j - \eta \nabla \calL ({\bm{\theta}^j_{p,t}}^\top z_{V_p}(\bm{x}_t^j), y_t^j), \label{eq:theta_update} \\
   w_{p,t+1}^j                 & = w_{p,t}^j \exp{-\eta \calL \lp \hat f^j_{p,t}(\bm{x}_t^j) \rp }, \label{eq:w_update}
\end{align}
where $\eta \in (0,1)$ is a learning rate.
The weights are normalized as $\bar w_{p,t}^j = w_{p,t}^j / \sum_{p=1}^P w_{p,t}^j$ to have
\begin{equation}
   \hat f^j_t (x) = \sum_{p = 1}^P \bar w_{p,t}^j \hat f^j_{p,t} (\bm{x}_t^j).
\end{equation}
The above equations represent how local online multi-kernel learning models can be built \cite{OMKL, Shen2019}.
In contrast to previous research, each node $j$ in our scenario can only communicate with its neighboring nodes in the set $\calS_j \subseteq V$.
Thus, the local model $\bm{\theta}^j_{p,t}$ of each node may be different from node to node.
Therefore, our algorithm must ensure that all nodes converge to the same model.
For this purpose, the local models are propagated using a gossip algorithm.
The nodes calculate the weighted average of the information provided by their neighbors at each communication round and use it to update their local information.
The weight Node $j$ assigns to the information coming from Node $i$ is defined as $w_{ij}'$.
Notice that the sum of weights $\sum_{i \in \calS_j} w_{ij}' = 1$ for all $j$ since the weights of a weighted average must add up to one.
We impose $w_{ij}' = w_{ji}'$ such that we can associate the weights to the edges of the undirected communication graph $G$.
Thus, the weights define the gossip matrix $\bm{W}'$, which is $J\times J$ and doubly stochastic
~\cite{xiao2004fast, boyd2004fastest, gharesifard2010does}.
The spectral gap of $\bm{W}'$ is denoted by $\rho = 1 - \lambda_2(\bm W') \in (0,1]$, where $\lambda_2(\bm W')$ represents the second eigenvalue of $\bm W'$ in descending order.

Such a gossip algorithm works very well when the nodes communicate their states perfectly with their neighbors. However, practically, only a few bits can be communicated, i.e., information needs to be quantized before being shared with the neighbors.
We use a random quantizer $Q: \R^n \to \R^n$, to represent an arbitrary vector $\bm{x} \in \R^n$ with $Q(\bm{x})$ in an efficient way.
Our results work for any random quantizer that satisfies
\begin{equation}
   \mathbb{E}_Q \lVert Q(\bm x) - \bm x\rVert^2 \leq (1 - \delta) \lVert \bm x\rVert^2, \quad \forall \bm x \in \R^n, \label{eq:Qbound}
\end{equation}
for some $\delta > 0$, which we will call the compression parameter.
Here, $\mathbb{E}_Q$ denotes the expected value with respect to the internal randomness of $Q(\cdot)$.
In simulations, we use the randomized quantizer of \cite{alistarh2017qsgd}.
Each element of a non-zero vector $\bm{v} \in \R^n$, i.e., $v_i$, is quantized by
\begin{equation}
   Q_M(v_i) = \lVert v \rVert \sign(v_i) \xi_i(\bm{v}, M),
\end{equation}
where $M = 2^b - 1$ is the number of quantization levels and defining $l = \lfloor M \tfrac{v_i}{\lVert v \rVert}  \rfloor$,
\begin{equation} \label{eq:quantizer}
   \xi_i(\bm{v}, M) =
   \begin{cases}
      \frac{l}{M}   & \text{with probability } 1 - M \tfrac{v_i}{\lVert v \rVert} + l, \\
      \frac{l+1}{M} & \text{otherwise},
   \end{cases}
\end{equation}
is represented by $b$ bits.
Following \cite[Lemma 3.1]{alistarh2017qsgd}, Eq. \eqref{eq:Qbound} holds for $\delta =1 -\min \lp \tfrac{2D}{M^2}, \tfrac{\sqrt{2D}}{M} \rp>0$.

\section{Algorithm and Regret Analysis}
We present \autoref{alg:gossipOMKL} for gossiped and quantized OMKL.
For ease of notation, we will denote $\calL({\bm{\theta}^j_{p,t}}^\top z_{V_p}(\bm{x}_t^j), y_t)$ as $\calL(\bm{\theta}^j_{p,t})$ from now on.
In \autoref{alg:gossipOMKL}, at each time instance, nodes collect their local data and transform them according to the RF approximation.
Then, the kernel losses are computed and used to update the kernel weights.
For the gossip step, we define a hidden state $\bm{h}_{p,t}^j \in \R^{2D}$ for each node $j$ that is known for all neighbors in $\calS_j$ because it is updated by the same quantized values known to all neighbors.
The new local state, $\bm{\theta}_{p,t}^j$,  will be a sum of $\bm{\theta}_{p,t-1}^{\prime j}$, which is an auxiliary variable where we store the local learning, plus a gossip step of size $\gamma$ with the information from the hidden states.
Subsequently, each node $j$ prepares the update to the hidden state by quantizing the difference between its local state $\bm{\theta}_{p,t}^j$ and the common hidden state $\bm{h}_{p,t}^j$.
This quantized difference is sent to the neighbors and is used by them to collectively update the hidden state.
Finally, each node performs local learning with a step size $\eta$, and stores it in the auxiliary variable $\bm{\theta}_{p,t}^{\prime j}$.

The role of a hidden state and quantized update is to have an accurate representation of the neighbor's states without the need for broadcasting the full state at each time instance. 

\begin{algorithm}[htbp] 
   \caption{Gossiped and Quantized OMKL at Node $j$}
   \begin{algorithmic}[1] \label{alg:gossipOMKL} 
      \STATE Initialize $w_p^j = 1/P$ and $\bm{h}_p^j = \bm{\theta}_p^j = \bm{\theta}_p^{\prime j} = 0$ for all $p$.
      \FOR {$t=1, \ldots, T$}
      \STATE Obtain data $\bm{x}_t^j$ and construct $z_p(\bm{x}_t^j)$ for all $p$.
      \STATE Compute $\calL(\hat f^j_{p,t} (\bm{x}_t^j))$ for all $p$.
      \STATE Update $w_{p,t}^j$ according to \eqref{eq:w_update}.
      \STATE Compute $\calL(\hat f^j_t(\bm{x}_t^j)) $
      \FOR {$p=1, \ldots, P$}
      \STATE Update using the hidden states $\bm{\theta}_{p,t}^j = \bm{\theta}_{p,t - 1}^{\prime j} + \gamma \sum_{i=1}^J w'_{ij} (\bm{h}_{p,t}^i - \bm{h}_{p,t}^j)$.
      \STATE Compress $\bm{q}_{p,t}^j = Q(\bm{\theta}_{p,t}^j - \bm{h}_{p,t}^j)$.
      \FOR {$i$ neighbor of $j$, including $j$}
      \STATE Send $\bm{q}_{p,t}^j$ to $i$ and receive $\bm{q}_{p,t}^i$.
      \STATE Update $\bm{h}_{p,t+1}^i = \bm{h}_{p,t}^i + \bm{q}_{p,t}^i$.
      \ENDFOR
      \STATE As in \eqref{eq:theta_update}, update $\bm{\theta}_{p,t}^{\prime j} = \bm{\theta}_{p,t}^j - \eta \nabla \calL (\bm{\theta}_{p,t}^j)$.
      \ENDFOR
      \ENDFOR
   \end{algorithmic}
\end{algorithm}
In what follows, we provide regret analysis to study the performance of our algorithm. First, we make the following assumptions similar to what is done in the literature.
\begin{assumption}\label{as:bounds_on_loss_gradient}
   Let us assume a $K$-smooth loss function
   \begin{equation}
      \lVert \nabla \calL(\bm{\theta}_1) - \nabla \calL(\bm{\theta}_2) \rVert \leq K \lVert \bm{\theta}_1- \bm{\theta}_2\rVert \quad \forall \bm{\theta}_1, \bm{\theta}_2 \in \R^{2D},
   \end{equation}
   with a bounded gradient $\lVert \nabla \calL (\bm{\theta}) \rVert \leq L$.
\end{assumption}

\begin{assumption}\label{as:convexity_loss}
   Let us assume a convex loss function  with respect to $\bm{\theta}$, i.e.,
   \begin{equation} \label{eq:convexity_loss}
      \calL(\bm{\theta}_{p,t}^j) - \calL(\bm{\theta}) \leq \nabla \calL(\bm{\theta}_{p,t}^j)^\top (\bm{\theta}_{p,t}^j -\bm{\theta}) \quad \forall \bm{\theta} \in \R^{2D}.
   \end{equation}
\end{assumption}

\begin{assumption} \label{as:connected_graph}
   Let us assume that we have a connected communication graph and a doubly stochastic gossip matrix $\bm{W}'$ whose entries are zero if and only if their corresponding edges are not present in the communication graph.
\end{assumption}

\begin{thm} \label{thm:main}
   Under Assumptions \ref{as:bounds_on_loss_gradient}, \ref{as:convexity_loss}, and \ref{as:connected_graph}, the sequences $\{\hat f^j_{p,t}\}$ and $\{\bar w _{p,t}\}$ generated by the gossiped and quantized OMKL satisfy
   \begin{multline} \label{eq:theoremeq}
      \sum_{j=1}^J \sum_{t=1}^{T} \expec{\calL \left( \sum_{p=1}^P \bar{w}_{p,t}^j \hat f^j_{p,t}(\bm{x}_t^j) \right)} - \sum_{j=1}^J \sum_{t=1}^{T}\calL \left( \hat f_{p}^*(\bm{x}_t^j) \right) \leq \\
      J\frac{\lVert \bm{\theta}^*_p \rVert^2}{2\eta} + \eta \frac{JTL^2}{2} + \eta \frac{JTL \sqrt{12}}{c} + \frac{J \ln P}{\eta} + \eta JT,
   \end{multline}
   where $\bm{\theta}^*_p$ is the optimal solution to the problem for the $p$-th kernel, $c = \frac{\rho^2\delta}{82}$, $\rho$ is the spectral gap of the gossip matrix $\bm W'$, and $\delta > 0$ is the quantizer's compression parameter.
\end{thm}
To prove \autoref{thm:main}, we first need to present \autoref{lemma:from_koloskova}, which bounds the difference between a single kernel loss and the optimal loss, and \autoref{lemma:kernel_weights_error}, which bounds the error added by the kernel weights.
\begin{lemma} \label{lemma:from_koloskova}
   Let us choose the consensus step size  as in \cite[Thm. 4.1]{koloskova2019decentralized}, i.e., $\gamma = \frac{\rho^2\delta}{16 \rho + \rho^2 + 4\beta^2 + 2\rho \beta^2 -8\rho \delta}$, where we have defined $\beta = \lVert \bm{I} - \bm{W'} \rVert_2 \in [0,2]$.
   Then, for $\eta \in (0,1]$ and under Assumptions \ref{as:bounds_on_loss_gradient}, \ref{as:convexity_loss}, and \ref{as:connected_graph}, we have
   \begin{multline}
      \sum_{j=1}^J \sum_{t=1}^{T} \expec{\calL \lp \hat f^j_{p,t}(\bm{x}_t^j) \rp} - \sum_{j=1}^J \sum_{t=1}^{T}  \calL \lp \hat f_{p}^*(\bm{x}_t^j) \rp  \leq \\
      J\frac{\lVert \bm{\theta}^*_p \rVert^2}{2\eta} + \eta \frac{JTL^2}{2} + \eta \frac{JTL \sqrt{12}}{c}.
   \end{multline}
\end{lemma}
\begin{proof}
   Similar to \cite{OMKL}, let us consider an arbitrary $\bm{\theta} \in \R^{2D}$.
   Define $\bar{\bm{\theta}}_{p,t} = \frac{1}{J} \sum_{j=1}^J \bm{\theta}_{p,t}^j$, and $\bar{\bm{\theta}}_{p,t}'$ and $\nabla \bar \calL (\bm{\theta}_{p,t})$ analogously.
   Since our gossip scheme \emph{preserves averages}, we can deduce $\bar{\bm{\theta}}_{p,t+1} = \bar{\bm{\theta}}_{p,t}' = \bar{\bm{\theta}}_{p,t} -\eta\nabla \bar \calL (\bm{\theta}_{p,t})$. Thus,
   \begin{align}
      \lVert \bar{\bm{\theta}}_{p,t+1} - \bm{\theta} \rVert^2 & = \lVert \bar{\bm{\theta}}_{p,t} -\eta\nabla \bar \calL (\bm{\theta}_{p,t})  - \bm{\theta} \rVert^2                     \\
                                                              & = \lVert \bar{\bm{\theta}}_{p,t}  - \bm{\theta} \rVert^2 + \eta^2 \lVert \nabla \bar \calL (\bm{\theta}_{p,t}) \rVert^2 \\
                                                              & -2\eta \nabla \bar \calL (\bm{\theta}_{p,t})^\top (\bar{\bm{\theta}}_{p,t} - \bm{\theta}). \label{eq:avg_minus_theta}
   \end{align}
   By \autoref{as:convexity_loss}, we have
   \begin{multline} \label{eq:convexity_extra}
      \calL (\bm{\theta}_{p,t}^j) - \calL(\bm{\theta}) \leq \nabla \calL(\bm{\theta}_{p,t}^j)^\top (\bm{\theta}_{p,t}^j + \bar{\bm{\theta}}_{p,t} - \bar{\bm{\theta}}_{p,t} - \bm{\theta}) \\
      \leq \nabla \calL(\bm{\theta}_{p,t}^j)^\top (\bar{\bm{\theta}}_{p,t}  - \bm{\theta}) + L \lVert \bm{\theta}_{p,t}^j - \bar{\bm{\theta}}_{p,t} \rVert.
   \end{multline}
   Inserting \eqref{eq:convexity_extra} into \eqref{eq:avg_minus_theta} and summing over $j$, we obtain
   \begin{multline} \label{eq:resultforj}
      \sum_{j=1}^J \lp \calL (\bm{\theta}_{p,t}^j) - \calL(\bm{\theta}) \rp \leq \frac{\eta}{2}J \lVert \nabla \bar \calL (\bm{\theta}_{p,t})\rVert^2 + \\
      J \frac{\lVert \bar{\bm{\theta}}_{p,t} - \bm{\theta} \rVert^2 - \lVert \bar{\bm{\theta}}_{p,t+1} - \bm{\theta} \rVert^2}{2\eta} + L\sum_{j=1}^J \lVert \bm{\theta}_{p,t}^j - \bar{\bm{\theta}}_{p,t} \rVert,
   \end{multline}
   where we have used \autoref{as:bounds_on_loss_gradient} to bound the gradient.
   Summing over $t$ and bounding again, we derive
   \begin{multline} \label{eq:resultforjandt}
      \sum_{t=1}^T \sum_{j=1}^J \lp \calL (\bm{\theta}_{p,t}^j) - \calL(\bm{\theta}) \rp\leq \frac{\eta}{2}J T L^2 + L\sum_{t=1}^T\sum_{j=1}^J \lVert \bm{\theta}_{p,t}^j - \bar{\bm{\theta}}_{p,t} \rVert\\
      + J \frac{\lVert \bar{\bm{\theta}}_{p,1} - \bm{\theta} \rVert^2 - \lVert \bar{\bm{\theta}}_{p,T+1} - \bm{\theta} \rVert^2}{2\eta}.
   \end{multline}
   Applying \cite[Lemma A.2]{koloskova2019decentralized} and Cauchy-Schwarz results in
   \begin{align}
      \sum_{j=1}^J \mathbb{E} \lVert \bm{\theta}_{p,t}^j - \bar{\bm{\theta}}_{p,t} \rVert^2 & \leq \eta^2 \frac{12 J L^2}{c^2} \implies               \\
      \sum_{j=1}^J \mathbb{E} \lVert \bm{\theta}_{p,t}^j - \bar{\bm{\theta}}_{p,t} \rVert   & \leq \eta \frac{JL \sqrt{12}}{c}. \label{eq:diff_bound}
   \end{align}
   The proof is complete, by using \eqref{eq:diff_bound} to bound \eqref{eq:resultforjandt} and choosing $\bm{\theta}_{p,1}^j = \bm 0$ and $\bm{\theta} = \bm{\theta}_p^*$.
\end{proof}

\begin{figure*}[htbp]
   \centering
   \includegraphics[width=.95\textwidth]{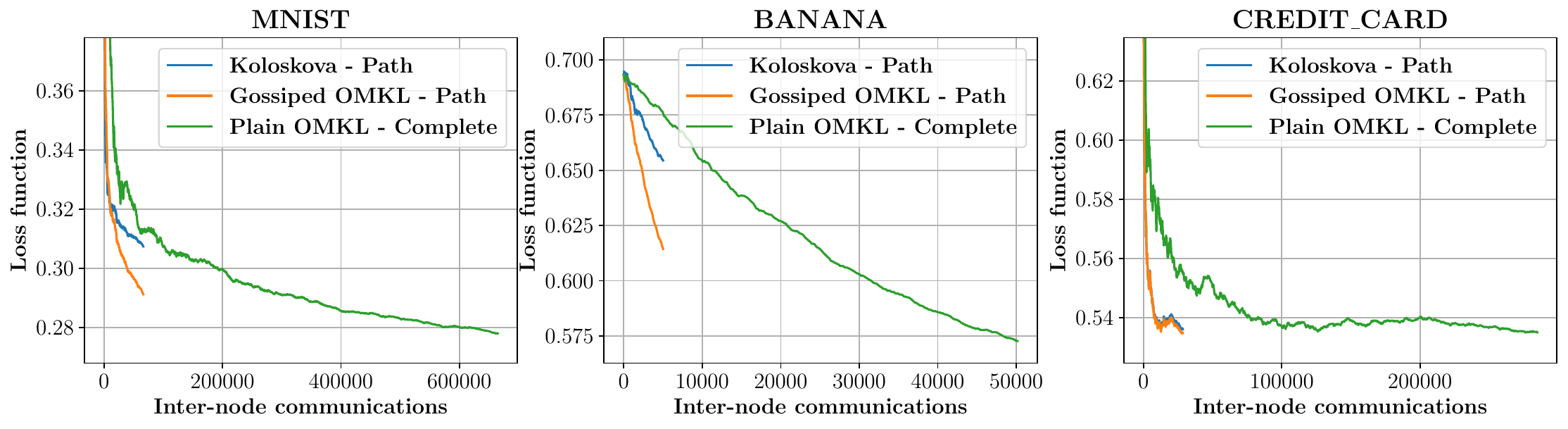}
   \caption{Comparison between the gossiped OMKL, OMKL from \cite{OMKL} and the single-kernel SGD, labelled Koloskova, from \cite{Koloskova2019}. We have used $\sigma = 1$ for the latter. The labels indicate the used topology for each algorithm.}
   \label{fig:LossComp}
\end{figure*}

\begin{lemma} \label{lemma:kernel_weights_error}
   Under the same conditions as \autoref{thm:main},
   \begin{multline}
      \sum_{j=1}^J \sum_{t=1}^{T} \sum_{p=1}^P \bar{w}_{p,t}^j \calL \lp  \hat f^j_{p,t}(\bm{x}_t^j) \rp - \sum_{j=1}^J \sum_{t=1}^{T} \calL \lp \hat f^j_{p',t}(\bm{x}_t^j) \rp \\
      \leq \frac{J \ln P}{\eta} + \eta JT.
   \end{multline}
\end{lemma}
The proof of \autoref{lemma:kernel_weights_error} follows the same steps as in \cite[Lemma 3]{OMKL}.

From the convexity of $\calL(\cdot)$,
\begin{equation}
   \calL \lp  \sum_{p=1}^P \bar{w}_{p,t} \hat f^j_{p,t}(\bm{x}_t) \rp \leq  \sum_{p=1}^P \bar{w}_{p,t} \calL \lp  \hat f^j_{p,t}(\bm{x}_t) \rp,
\end{equation}
and \autoref{thm:main} follows immediately using \autoref{lemma:from_koloskova} and \autoref{lemma:kernel_weights_error}.

\section{Experimental results and conclusions} \label{sec:experiments}
To validate \autoref{thm:main}, we have tested \autoref{alg:gossipOMKL}
with real datasets for binary classification, with different topologies, and using different values of quantization level.
We have used the well-known Kernel Logistic Regression (KLR) loss function \cite{scholkopf2002learning}, i.e., the loss function \eqref{eq:loss_function} is defined as
\begin{equation}\label{eq:klr_loss}
   \ln \lp 1 + \exp(-y \cdot \bm{\theta}^\top z_V(\bm{x})) \rp + \lambda \lVert \bm{\theta} \rVert^2,
\end{equation}
where we have used $\Omega(\lVert \bm{\theta} \rVert^2) = \lVert \bm{\theta} \rVert^2$.
We have conducted our experiments with three datasets: Banana, Credit-Card, and MNIST.
The synthetic data from the Banana dataset \cite{ratsch2001soft} ($n = 5300, d = 2$) are two non-linearly separable clusters.
The Credit-Card dataset \cite{yeh2009comparisons} ($n = 30000, d = 2$) contains data from credit card users, such as their credit history and whether they have defaulted or not.
It is obtained from the UCI machine learning repository \cite{Dua:2019}.
The MNIST dataset \cite{lecun1998mnist} ($n = 70000, d = 784$) is a labeled set of handwritten digits from 0 to 9.
We divide them into two classes, those that are number 8 and those that are not.

Our experimental setup has $J=20$ nodes, dimension $D = 20$ for our RF approximation, regularization parameter $\lambda = 0.001$, and three Gaussian kernels with $\sigma \in \{1, 3, 5\}$, where a Gaussian kernel with parameter $\sigma$ is defined as $\kappa(\bm{x}, \bm{x'}) = \exp{[-\lVert \bm{x} - \bm{x'}\rVert^2 / (2\sigma^2)]}$.
Simulations have been performed 100 times with different sets of Random Features and the corresponding mean of the loss function is plotted.
We report the loss instead of the classification accuracy because the difference in the latter is minimal.
We have used the quantizer described in \eqref{eq:quantizer} with 7 levels of quantization, that is, 3 bits per element in any transmitted array. Our learning rates are $\eta = 0.01$ and $\gamma = 0.9 \eta=0.009$.

\autoref{fig:LossComp} compares the performance of our algorithm versus two benchmarks.
Since \autoref{alg:gossipOMKL} can be viewed as an extension of \cite{OMKL} to non-complete graphs, the first benchmark for comparison is the OMKL algorithm from \cite{OMKL}.
There is a small difference between the performance of OMKL and that of our gossiped OMKL although OMKL requires many more inter-node communications.
This is despite the fact that OMKL is run over a complete topology, but the gossiped OMKL is executed over the worst-case scenario, i.e., a path topology that includes much smaller number of connections and as a result requires much less communication. 

Gossiped OMKL can also be considered as an extension of \cite{Koloskova2019} to multi-kernel learning, which justifies using the single-kernel SGD from \cite{Koloskova2019} as our second benchmark.
Gossiped OMKL using three kernels clearly outperforms the single-kernel approach, named Koloskova in \autoref{fig:LossComp}.

\begin{figure}[htbp]
   \centering
   \includegraphics[width=.9\linewidth]{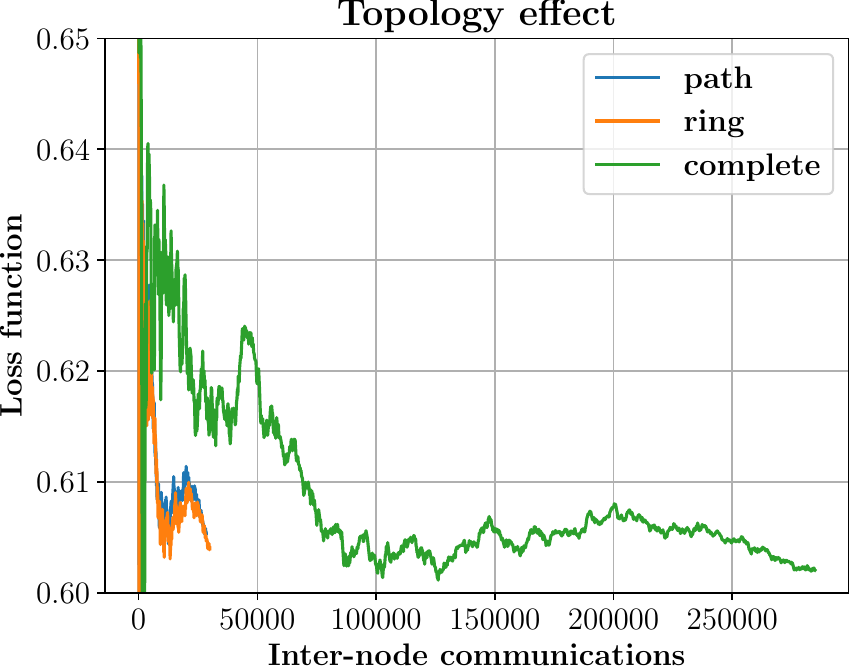}
   \caption{The effect of graph topology, without quantization. Average loss function at each iteration, with the Credit-Card dataset. Since the Credit-Card dataset has $n = 30000$ data observations, and $J=20$ nodes, we have run the algorithm $1500$ iterations.}
   \label{fig:topolgy}
\end{figure}

Although not shown in the figure, we have observed that, for values of $\eta$ and $\gamma$ chosen in \autoref{fig:LossComp}, or smaller values, the choice of topology does not affect the performance of our algorithm. To observe the effect of topology on algorithm performance, larger step sizes ($\eta = 0.1$ and $\gamma = 0.09$) are chosen in Fig. \ref{fig:topolgy}.
The figure shows simulations using the Credit-Card dataset and without quantization on three different topologies: the complete graph, the ring, and the path.
We can observe that more densely connected communication graphs lead to better performance.
We have also tested the algorithm for $M \geq 7$ levels of quantization, that satisfies the condition $\delta > 0$ in \eqref{eq:quantizer}, and all of them perform as well as the non-quantized version, with the loss function differences in the order of $10^{-6}$.

The results show that our gossiped OMKL algorithm can successfully extend the OMKL algorithm \cite{OMKL} to non-complete graphs and the distributed learning algorithm in \cite{Koloskova2019} to multi-kernel learning.

\bibliographystyle{IEEEtran}
\bibliography{references}

\end{document}